%% file: main.tex
\documentclass{ecai} 
\usepackage{latexsym}
\usepackage{amssymb}
\usepackage{amsmath}
\usepackage{multirow}
\usepackage{amsthm}
\usepackage{booktabs}
\usepackage{enumitem}
\usepackage{graphicx}
\usepackage{color}
\usepackage{bm}
\usepackage{caption}
\usepackage{xspace}
\usepackage{pifont}
\usepackage{float}   
\usepackage[table,xcdraw]{xcolor}
\usepackage{booktabs} 
\usepackage{array}  
\usepackage{adjustbox} 

\definecolor{lighterblue}{rgb}{0.8, 0.9, 1.0}
\definecolor{lightblue}{rgb}{0.678, 0.847, 0.902}
\definecolor{gold}{HTML}{FFD700}
\definecolor{blue}{RGB}{30,144,255} 
\definecolor{red}{RGB}{255,99,71}  

\newcommand{\BibTeX}{B\kern-.05em{\sc i\kern-.025em b}\kern-.08em\TeX}

\DeclareRobustCommand\onedot{\futurelet\@let@token\@onedot}
\def\eg{\emph{e.g., }}

\newcommand{\mparagraph}[1]{\vspace{1mm}\noindent{\textbf{#1.}\hspace{1mm}}}

\begin{document}


\begin{frontmatter}


\paperid{2394} 




\title{CorrMoE: Mixture of Experts with De-stylization Learning for Cross-Scene and Cross-Domain Correspondence Pruning}


\author[A]{\fnms{Peiwen}~\snm{Xia}}
\author[A]{\fnms{Tangfei}~\snm{Liao}}
\author[B]{\fnms{Wei}~\snm{Zhu}} 
\author[B]{\fnms{Danhuai}~\snm{Zhao}} 
\author[B]{\fnms{Jianjun}~\snm{Ke}} 
\author[C]{\fnms{Kaihao}~\snm{Zhang}} 
\author[A]{\fnms{Tong}~\snm{Lu}}
\author[A]{\fnms{Tao}~\snm{Wang}}

\address[A]{Nanjing University}
\address[B]{China Mobile Zijin Innovation Institute}
\address[C]{Harbin Institute of Technology (Shenzhen)}

\begin{abstract}
Establishing reliable correspondences between image pairs is a fundamental task in computer vision, underpinning applications such as 3D reconstruction and visual localization. Although recent methods have made progress in pruning outliers from dense correspondence sets, they often hypothesize consistent visual domains and overlook the challenges posed by diverse scene structures. In this paper, we propose CorrMoE, a novel correspondence pruning framework that enhances robustness under cross-domain and cross-scene variations. To address domain shift, we introduce a De-stylization Dual Branch, performing style mixing on both implicit and explicit graph features to mitigate the adverse influence of domain-specific representations. For scene diversity, we design a Bi-Fusion Mixture of Experts module that adaptively integrates multi-perspective features through linear-complexity attention and dynamic expert routing. Extensive experiments on benchmark datasets demonstrate that CorrMoE achieves superior accuracy and generalization compared to state-of-the-art methods. The code and pre-trained models are available at \url{https://github.com/peiwenxia/CorrMoE}.
\end{abstract}

\end{frontmatter}


\begin{figure}[t]
\begin{center}
    \includegraphics[width=1.\linewidth]{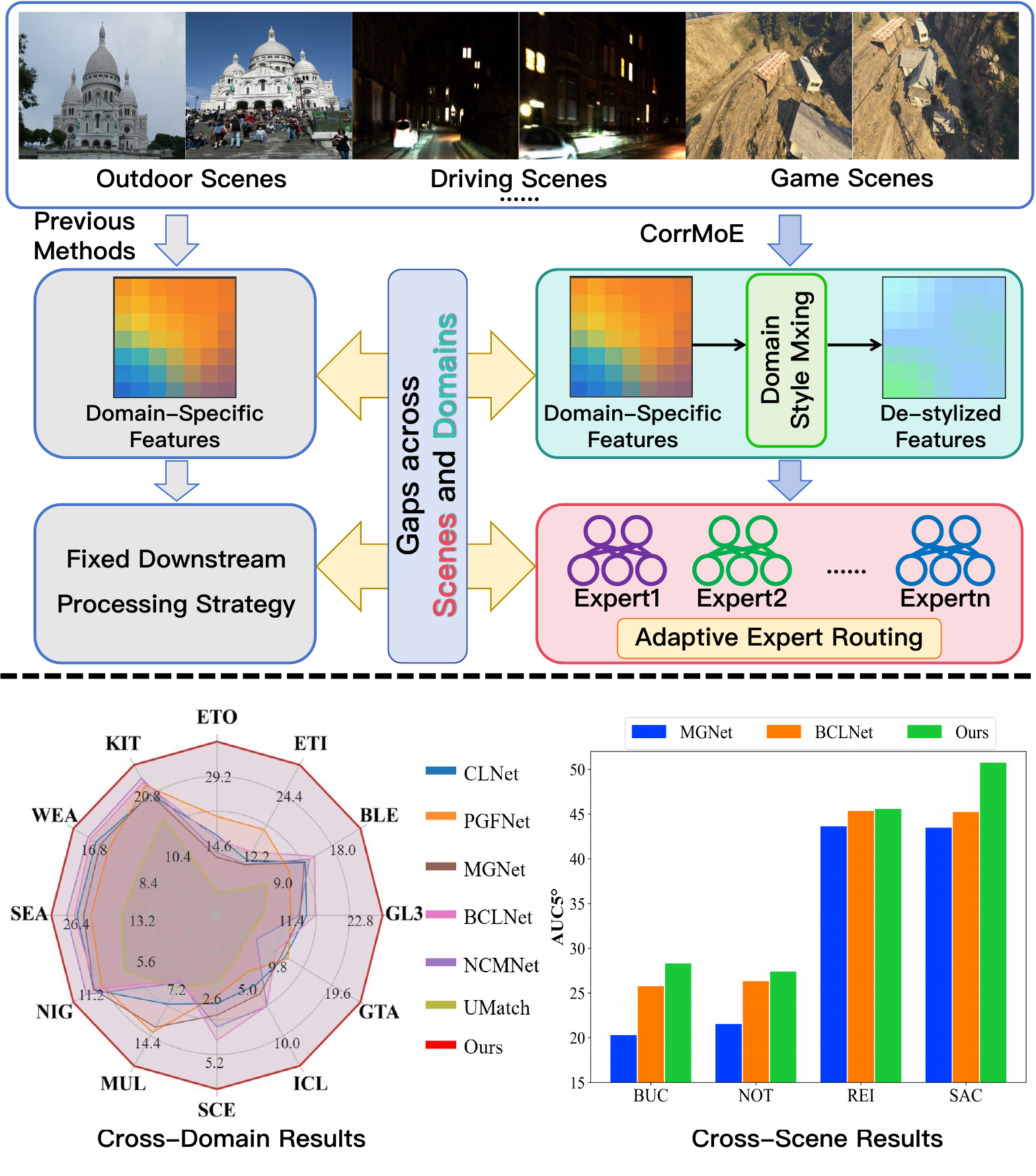}
    \caption{Comparison between previous methods and ours. \textbf{Top}: Facing the challenges of different scenes and domains, CorrMoE first de-stylizes the domain-specific local graph features, then applies a multi-expert approach to handle different scenes. \textbf{Bottom}: In terms of the AUC@5° metric, CorrMoE outperforms recent SOTA methods such as BCLNet~\cite{miao2024bclnet} and MGNet~\cite{dai2024mgnet} in both cross-domain and cross-scene settings. Cross-domain evaluations are conducted using 12 datasets from the Zero Shot Evaluation Benchmark~\cite{xuelun2024gim}, while cross-scene evaluations are conducted using 4 scenes split from the YFCC100M test set~\cite{Thomee2016}.}\label{fig:page1}
\end{center}
\end{figure} 

\section{Introduction}
\label{sec:intro}
Accurate two-view correspondences are crucial for a wide range of computer vision and robotics tasks, \eg visual localization~\cite{sarlin2021back}, stereo matching~\cite{yao2021decomposition}, simultaneous localization and mapping~\cite{campos2021orb, mur2017orb, mur2015orb}, and 3D reconstruction~\cite{agarwal2011building, kerbl3Dgaussians, schonberger2016structure}. Typically, putative correspondences are generated using the off-the-shelf detector-descriptors~\cite{DeTone2018,Lowe2004}. However, due to challenging real-world variations between images, such as rotations, viewpoint changes, illumination shifts, and occlusions, these initial correspondences often contain a significant proportion of false correspondences. To reduce the negative impact of these mismatches on downstream tasks, correspondence pruning methods, which aim to identify and retain geometrically consistent correspondences, have gained considerable attention.

Traditional correspondence pruning methods, like RANSAC~\cite{Fischler1981} and its variants~\cite{ Barath2019,Chum2005a}, use a hypothesize-and-verify strategy to maximize inliers. While effective in specific cases, their execution time grows exponentially with the number of outliers, making them inefficient in scenarios with high outlier ratios. These limitations have prompted the rise of learning-based approaches, which offer promising alternatives by leveraging data-driven models to handle outliers more effectively.

Learning-based correspondence pruning methods, inspired by LFGC~\cite{Yi2018}, typically focus on correspondence classification and camera pose estimation. CLNet~\cite{Zhao2021}, VSFormer~\cite{liao2023vsformer}, and NCMNet~\cite{liu2023ncm} use the K-Nearest Neighbor (KNN) algorithm to connect neighboring correspondences in an explicit way. In contrast, OANet~\cite{Zhang2019}, U-Match~\cite{li2024u}, and BCLNet~\cite{miao2024bclnet} capture consistent neighbors implicitly by modulating the local context between the pooling and unpooling operations. 
To better leverage the contextual features extracted from different types of graphs, recent advances~\cite{dai2024mgnet,liao2023sga} concatenate implicit and explicit graphs and builds the global graph based on them.

Existing methods~\cite{dai2024mgnet,liao2023sga,liu2023ncm,miao2024bclnet} capture contextual features from implicit and explicit aspects, which are subsequently passed to fixed downstream modules for inlier prediction. However, these methods fail to fully process these features from a cross-domain and cross-scene perspective, limiting their applicability. On the one hand, extracted contextual features are often tied to specific domain characteristics, making existing pruning methods prone to significant performance degradation when the domain shifts in real-world applications. On the other hand, these methods fail to consider the diverse distributions of putative sets, making them effective for common scenes but less capable with rare or complex ones.

To address these challenges, we introduce CorrMoE, a novel framework that effectively combines MoE with de-stylization learning, as illustrated in Fig.~\ref{fig:page1}. The core components of CorrMoE include the De-stylization Dual Branch and the Bi-Fusion Mixture of Experts (MoE). The De-stylization Dual Branch is designed to simulate cross-domain features within a single domain, which incorporates the Progressive Mixstyle Module, the implicit and explicit branches. Specifically, the Progressive Mixstyle Module mixes implicit and explicit graph styles. The implicit branch abstracts and aggregates local graph features via differentiable pooling and its reverse operation, while the explicit branch leverages KNN graphs and multi-dimensional attention mechanisms to establish relationships between neighboring correspondences. Following this, the Bi-Fusion MoE facilitates the further fusion of graphs and enhances the model's adaptability to diverse scenes. To be specific, this module employs a linear-complexity attention mechanism to strengthen the interaction between graph features. Additionally, it introduces a dynamic expert routing mechanism that enables the model to focus on graph nodes from multiple perspectives, allowing for precise capture of scene-specific features. As demonstrated in Fig.~\ref{fig:page1}, our CorrMoE effectively addresses the challenges of different tasks, surpassing existing methods in both cross-scene and cross-domain settings across multiple benchmark evaluations.

The contributions of this paper are three-fold: (1) We propose CorrMoE, a MoE-based framework incorporating de-stylization learning for the correspondence pruning task, ensuring robust performance across diverse domains and scenes. To our knowledge, CorrMoE is the first framework to introduce the MoE mechanism for addressing the two-view correspondence pruning problem. (2) We introduce two key modules: the De-stylization Dual Branch and the Bi-Fusion MoE Module. The De-stylization Dual Branch mitigates domain biases by mixing the local graph styles implicitly and explicitly, providing a more robust representation for pruning. The Bi-Fusion MoE Module integrates diverse contexts through linear-complexity attention and dynamic expert routing, improving the model’s ability to handle complex and rare scenes. (3) Through extensive experiments on benchmark datasets, we demonstrate that CorrMoE outperforms state-of-the-art methods in terms of adaptability and generalization, establishing a strong foundation for correspondence pruning in real-world scenarios.

\section{Related Work}
\label{sec:related_work}

\subsection{Correspondence Pruning}
Traditional method RANSAC~\cite{Fischler1981} and its variants (\textit{e.g.}, USAC~\cite{Raguram2012}, MAGSAC~\cite{Barath2019}, MLESAC~\cite{torr2000mlesac}, PROSAC~\cite{Chum2005a}) are initially used for geometric model estimation via iterative sampling. They hypothesize a model from a minimal subset of correspondences and verify it by counting inliers. However, RANSAC-based methods are often task-specific and struggle with larger datasets and high outlier ratios, which leads to unmanageable growth in computational time.

The rise of deep learning has made significant advancements in correspondence pruning~\cite{Yi2018, Zhang2019, Zhao2021, Dai2022, liu2023ncm, liao2023vsformer, dai2024mgnet, miao2024bclnet, zhu2024corradaptor}. One of the pioneering approaches, LGFC~\cite{Yi2018}, considers correspondence pruning as a combination of a binary classification task and an essential matrix regression task. The following works have focused on designing explicit and implicit modules to capture more complex contexts. Notably, OANet~\cite{Zhang2019} obtains local context implicitly by mapping correspondences to clusters via a differential pooling operation. CLNet~\cite{Zhao2021} derives consensus scores explicitly using dynamic graphs from local to global. Recently, MGNet~\cite{dai2024mgnet} proposes Graph Soft Degree Attention to fully leverage sparse correspondence information in the global graph. BCLNet~\cite{miao2024bclnet} utilizes a parallel context learning strategy to model interactions between different contexts. Although the existing learning-based methods perform well on in-domain datasets, their generalization ability on out-of-domain datasets is still constrained. To address this problem, we introduce a de-stylization network architecture, enabling better adaptation to cross-scene and cross-domain settings.

\subsection{Mixture-of-Experts Models}

Adopting a divide-and-conquer strategy, the MoE models~\cite{jordan1994hierarchical, masoudnia2014mixture} split the problem space among multiple experts and adaptively merge their separate results. The earliest variant of MoEs activates all experts densely for the input, resulting in a high computational burden~\cite{eigen2013learning}. By assigning varying weights to each expert, the sparsely-gated MoE~\cite{shazeer2017outrageously} achieves significant improvements in overall performance and lays the foundation for future advancements in the field. With the efficient routing mechanisms, models such as~\cite{fedus2022switch,lepikhin2020gshard} improve capacity while keeping computational costs low. Recently, MoE layers have been extensively utilized in natural language processing, frequently incorporated into key structures such as the FFN layer of transformers. In the computer vision area, MoE layers are applied to various tasks. For instance, previous works have employed experts with a shared decoder for classification tasks~\cite{gross2017hard}, introduced a tailored mixture of adapters for multi-source adaptive image fusion~\cite{zhu2024task}, and selected optimal experts to facilitate prompt fusion of adaptive visual features in 3D vision~\cite{ma20253d}. This paper introduces a correspondence pruning method based on the MoE mechanism for the first time.

\begin{figure*}[t]
\begin{center}
    \includegraphics[width=0.9\linewidth]{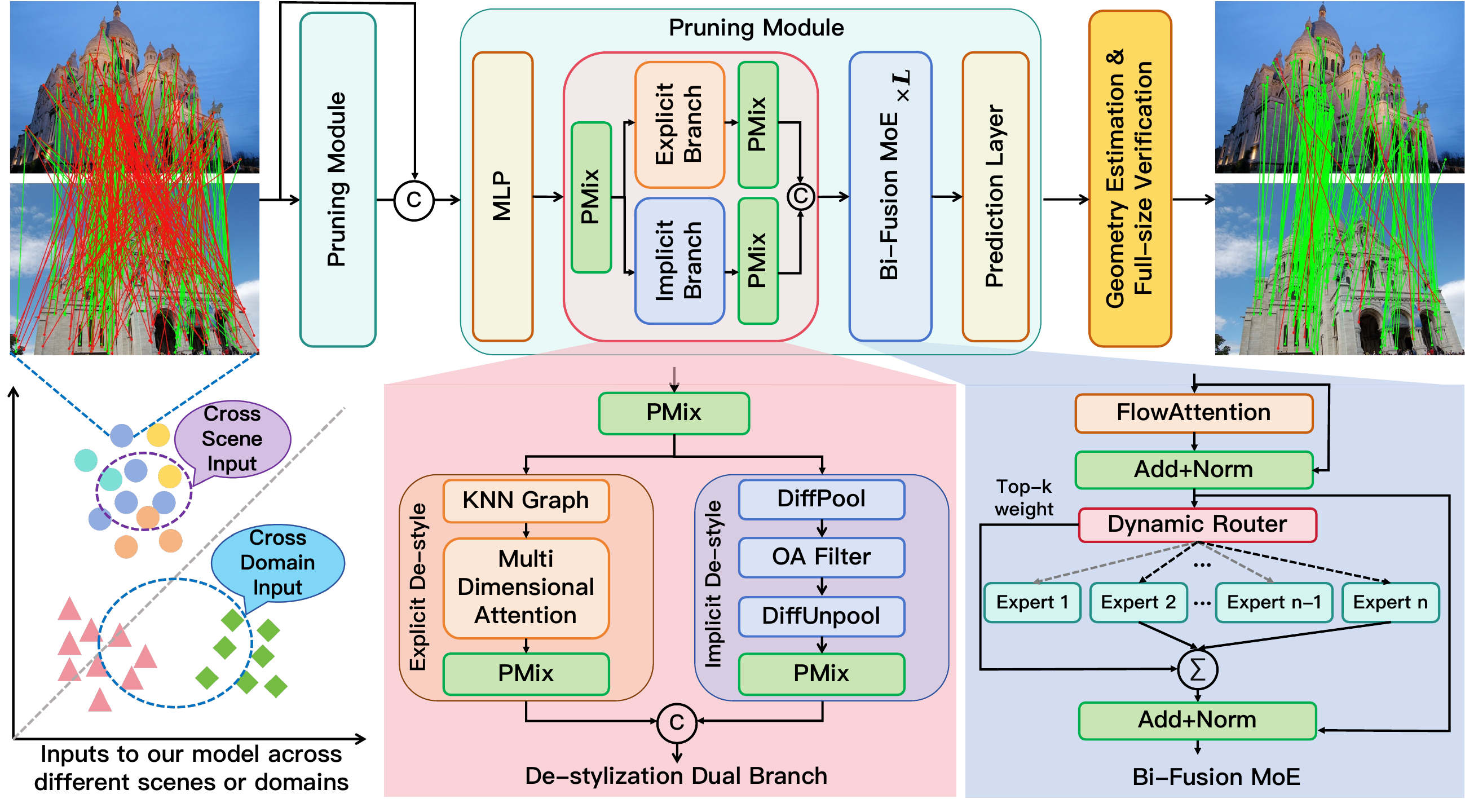}
    \caption{\textbf{Overview of our proposed CorrMoE, a framework that addresses both cross-domain and cross-scene correspondence pruning challenges. In our CorrMoE, the De-stylization Dual Branch facilitates cross-domain generalization, while the Bi-Fusion MoE enhances cross-scene adaptation. Initial correspondence sets with diverse domain styles or scene variations are progressively aligned and verified through the pruning module twice.}}\label{CorrMoE}
\end{center}
\end{figure*} 

\section{Methodology}
\label{sec:method}

\subsection{Problem Formulation}
Given a pair of images $(\mathbf{I}, \mathbf{I'})$, we utilize existing feature detectors~\cite{Lowe2004, DeTone2018} followed by a nearest neighbor matcher to build the initial correspondence set $\bm{C}$,
\begin{equation}
\bm{C} = \{ \bm{c}_1, \bm{c}_2, \bm{c}_3, \ldots, \bm{c}_N \} \in \mathbb{R}^{N \times 4}, \quad \bm{c}_i = (x_i, y_i, x'_i, y'_i),
\end{equation}
where $\bm{c}_i$ is a correspondence between keypoints $(x_i, y_i)$ in image $\mathbf{I}$ and $(x'_i, y'_i)$ in image $\mathbf{I'}$, which are normalized under camera intrinsics.

With $\bm{C}$ as input, Correspondence pruning aims to filter out incorrect matches from the initial correspondence set and accurately recover camera poses. As shown in Fig.~\ref{CorrMoE}, following~\cite{Zhao2021}, we adopt an iterative framework that employs the pruning module twice for correspondence pruning. In this module, the input sequentially passes through the perceptron layer, the De-stylization Dual Branch, the Bi-Fusion MoE, and the prediction layer. This process produces a final set of weights $\bm{\hat{W}}_2 = \{ \bm{w_1}, \ldots, \bm{w_i}, \ldots, \bm{w_{N'}} \}$, where each $\bm{w_i}$ represents the likelihood that the corresponding candidate is an inlier. The above process can be expressed as,
\begin{align}
(\bm{\hat{C}}_1, \bm{\hat{W}}_1) &= f_{\theta}(\bm{C}), \\
(\bm{\hat{C}}_2, \bm{\hat{W}}_2) &= f_{\phi}(\bm{\hat{C}}_1, \bm{\hat{W}}_1),
\end{align}
where $\bm{\hat{C}_2}$ and $\bm{\hat{W}}_2$ denote the final candidate set and the weights
of the final candidates. $f_{\theta}(\cdot)$ and $f_{\phi}(\cdot)$ denote the first and second iterations with learnable parameters $\theta$ and $\phi$, respectively.
Then a weighted eight-point algorithm $g(\cdot,\cdot)$ is employed to regress the essential matrix $\bm{\hat{E}}$, which can be formulated as,
\begin{align}
\bm{\hat{E}} = g(\bm{\hat{C}_2}, \bm{\hat{W}}_2).
\end{align}
Finally, we adopt the full-size verification operation to verify $\bm{\hat{E}}$ on the correspondence set $\bm{C}$ and retrieve the inliers that are mistakenly discarded during the correspondence pruning process. This process can be formulated as,
\begin{align}
\bm{\hat{D}} = h(\bm{\hat{E}}, \bm{C}),
\end{align}
where $h$ denotes the verification operation, and $\bm{\hat{D}}$ denotes the predicted symmetric epipolar distance set.

\subsection{De-stylization Dual Branch}
Existing methods~\cite{liao2023sga, liu2023ncm, dai2024mgnet, miao2024bclnet} based on local graph representations excel in in-domain generalization but often fail in out-of-domain scenes. This problem arises from the over-introduction of stylized features in the original correspondence set. To overcome this, we propose a De-stylization Dual Branch that simulates cross-domain style variations in implicit and explicit graphs, improving model adaptability and robustness. As shown in Fig.~\ref{CorrMoE}, the input is first processed by Progressive Mixstyle, followed by parallel explicit and implicit branches. Then, their outputs are concatenated to form the output. The details of the De-stylization Dual Branch are as follows.

\mparagraph{Progressive Mixstyle}
The De-stylization Dual Branch starts with the proposed Progressive Mixstyle (PMix), an improved version of Mixstyle~\cite{zhou2021domain}, designed for a more domain-invariant representation. Specifically, in contrast to the original Mixstyle with fixed application probability, the probability of using PMix progressively increases during the training, which is denoted as,
\begin{equation}
p_{mix}=p_{mix}^{start}+\delta(p_{mix}^{end}-p_{mix}^{start}),
\end{equation}
where $\delta=t/T$, with $t$ being the current epoch and $T$ the total number of training epochs. $p_{mix}^{start}$ and $p_{mix}^{end}$ are the probabilities at the start and end of training, respectively. When $p_{mix}$ is small, module PMix tends to be skipped, and the model focuses on the core patterns in-domain, thereby stabilizing the early stages of training. As $p_{mix}$ increases, the model is exposed to more diverse styles introduced by PMix, enhancing its generalization across domains.

In the PMix module, given an input batch $\bm{F}$, we first obtain $\tilde{\bm{F}}$ by randomly shuffling $\bm{F}$ along the batch dimension. Then, $\gamma_{mix}$ and $\beta_{mix}$ is calculated, where $\gamma_{mix}$ is a weighted combination of the standard deviations of $\bm{F}$ and $\tilde{\bm{F}}$, and $\beta_{mix}$ is a weighted combination of the means of $\bm{F}$ and $\tilde{\bm{F}}$. This process can be formulated as,
\begin{align}
\bm{\gamma}_{mix} &= \lambda\sigma(\bm{F}) + (1-\lambda)\sigma(\tilde{\bm{F}}), \\
\bm{\beta}_{mix} &= \lambda\mu(\bm{F}) + (1-\lambda)\mu(\tilde{\bm{F}}),
\end{align}
where $\lambda$ denotes an instance-wise weight, $\lambda \sim Beta(\alpha,\alpha)$ with $\alpha$ being a hyper-parameter. $\sigma(\cdot)$ and $\mu(\cdot)$ represent the operation of calculating the standard deviation and the mean of input along the channel dimension, respectively. Then, we employ $\bm{\gamma}_{mix}$ and $\bm{\beta}_{mix}$ to styled-normalized $\bm{F}$ and obtain $\bm{F}_{mix}$, which is formulated as,
\begin{equation}
\bm{F}_{mix} = \bm{\gamma}_{mix}\frac{\bm{F} - \mu(\bm{F})}{\sigma(\bm{F})} + \bm{\beta}_{mix}.
\end{equation}
After the PMix module, $\bm{F}_{mix}$ is fed into the dual branch to capture local graph features implicitly and explicitly.

\mparagraph{Implicit Graph Branch}
The Implicit Graph Branch utilizes the Order-Aware Block~\cite{Zhang2019} to exploit neighbors and encode the local context in an implicit way. The Order-Aware Block consists of the DiffPool operation, the OA Filter operation, and the DiffUnpool operation. Given a feature set $\bm{F}_{mix} = \{ f_i \}_{i=1}^N \in \mathbb{R}^{C \times N \times 1}$, the DiffPool operation is used to map $\bm{F}_{mix}$ into a coarse-grained graph $\mathcal{\bm{G}}^I \in \mathbb{R}^{C \times M \times 1}$. Then we use the OA Filter operation to encode coarse-grained graphs $\mathcal{\bm{G}}^I$, so that the local representation is enriched. After that, the DiffUnpool operation is applied to restore the graph to its original size. Finally, the PMix module is applied to the graph feature. The above process can be formulated as,
\begin{align}
\mathcal{\bm{G}}^I \hspace{7pt} &= \text{DiffPool}(\bm{F}_{mix}),\\
\bm{F}^I \hspace{7pt} &= \text{DiffUnpool}(\bm{F}_{mix}, \text{OA}(\mathcal{\bm{G}}^I)), \\
\bm{F}^{I}_{mix} &= \text{PMix}(\bm{F}^I),
\end{align}
where $\bm{F}^I$ denotes the implicit local
graph feature set, $\text{PMix}(\cdot)$ refers to the PMix module, and $\bm{F}^{I}_{mix}$ is the final output of the implicit branch.

\mparagraph{Explicit Graph Branch}
The explicit graph branch begins by constructing a KNN-based graph using the Euclidean distances between each correspondence:
\begin{equation}
\mathcal{\bm{G}}^E_i=\{\bm{V}^E_i,\bm{E}^E_i\},i \in [1, N],
\end{equation}
where $\bm{V}^E_i=\{f_{ij}\}^k_{j=1}$ denotes the neighbor set of $\bm{c}_i$ in the feature space. $\bm{E}^E_i=\{\bm{e}_{ij}\}^k_{j=1}$ is constructed by concatenating the selected correspondence feature $f_i$ and the residual value with its $k$-nearest neighbors that is expressed as,
\begin{equation}
\bm{e}_{ij} = [f_i, f_{ij}],
\end{equation}
where $[\cdot, \cdot]$ denotes the concatenation operation along channel dimension.

To further explore the rich contextual information contained in the explicit graph, inspired by~\cite{liao2023vsformer}, the Multi-Dimensional Attention Block is applied to sequentially capture interactions along spatial, neighbor, and channel dimensions. As the attention mechanisms are similar across these dimensions, we detail only the channel attention for brevity.

In the process of channel attention operation, the explicit graph feature is first encoded by PointCN~\cite{Yi2018}. Then, the neighbor dimension of the graph feature is squeezed by element-wise adding the results of the MaxPool and AvgPool operations. Subsequently, a gating mechanism is applied to derive attention scores. Finally, these attention scores are element-wise multiplied with the global graph feature, adding a residual to produce the output. The above operations can be expressed as,
\begin{align}
\mathcal{\bm{G}}^{E'} &= \text{PointCN}(\mathcal{\bm{G}}^{E}), \\
\bm{A}_{ca} &= \text{Gate}(\text{MaxPool}(\mathcal{\bm{G}}^{E'})+\text{AvgPool}(\mathcal{\bm{G}}^{E'})), \\
\mathcal{\bm{G}}^{E}_{ca} &= \mathcal{\bm{G}}^{E'} \odot \bm{A}_{ca} + \mathcal{\bm{G}}^{E'},
\end{align}
where $\text{Gate}(\cdot)$ consists of an MLP layer and a sigmoid function.

Similar to channel attention, neighbor and spatial attention are also applied in the Multi-Dimensional Attention Block. Next, the AnnualConv operation proposed by~\cite{Zhao2021} aggregates the explicit graph feature, followed by the PMix module. The whole process can be described as:
\begin{align}
\mathcal{\bm{G}}^E_{attn} &= \text{CA}(\text{NA}(\text{SA}(\mathcal{\bm{G}}^E))), \\
\mathcal{\bm{G}}^E_{aggr} &= \text{AnnualConv}(\mathcal{\bm{G}}^E_{attn}), \\
\mathcal{\bm{G}}^E_{mix} &= \text{PMix}(\mathcal{\bm{G}}^E_{aggr}),
\end{align}
where CA, NA, and SA represent channel, neighbor, and spatial attention operation, respectively.

\subsection{Bi-Fusion MoE Module}
To enhance the interaction between the explicit and implicit graph features extracted by the De-stylization Dual Branch, we propose a Bi-Fusion MoE Module. This module facilitates the model in capturing both common and unique cross-scene structures from the perspectives of multiple experts, each focusing on different local geometric or visual patterns, thereby enhancing its dynamic adaptability when dealing with sparse or unfamiliar scenarios.

First, an attention mechanism along with residual connections and normalization is applied to realize the initial fusion of local graph features $\bm{F}^{IE}=[\bm{F}^I_{mix},\mathcal{\bm{G}}^E_{mix}]$. Traditional attention mechanisms suffer from quadratic complexity with respect to the input size, which can lead to significant computational overhead, especially when processing large graph features. To overcome this problem, we adopt FlowAttention~\cite{wu2022flowformer}, a linear complexity alternative that decomposes the attention process into three components: competition, aggregation, and allocation. All three operations are based on efficient broadcasting, summation, and element-wise multiplication rather than explicit attention matrices. As a result, FlowAttention not only ensures scalability to long sequences but also maintains strong representational performance. The above process can be formulated as,
\begin{align}
\bm{F}^{IE}_{attn}&=\text{FlowAttn}(\bm{F}^{IE}),\\
\bm{F}^{IE'}_{attn}&=\text{Norm}(\bm{F}^{IE}+\bm{F}^{IE}_{attn}).
\end{align}
Next, inspired by \cite{fedus2022switch,shazeer2017outrageously}, we construct the MoE layer, which consists of a routing gate and a set of customized modulation experts $\{E_1, E_2, \cdots, E_n\}$. Guided by a routing prompt, the gate dynamically assigns weights, directing the features to the most suitable experts for targeted modulation and processing. The graph feature is fed into the gate network to generate weights as follows:
\begin{align}
\bm{Z}&=\text{MLP}(\bm{F}^{IE'}_{attn}),\\
\bm{W}&=\text{Softmax}(\text{TopK}(\bm{Z})).
\end{align}
This process transforms the input into routing logits to guide expert selection. Using the TopK algorithm~\cite{shazeer2017outrageously}, experts with the highest weights are selected, while the remaining weights are set to negative infinity. This process can be formulated as:
\begin{align}
\bm{F}^{IE}_{fuse}=\sum_{i=1}^N \bm{W}_i \cdot \bm{E}_i(\bm{F}^{IE'}_{attn}),
\end{align}
where $\bm{E}_i(\cdot)$ denotes the feature projection operation of the $i$-th expert, $\bm{W}_i$ denotes the weight assigned to the $i$-th expert. Finally, after another residual connection and normalization, we obtain the output of the Bi-Fusion MoE module.

\subsection{Loss Function}
Following the existing methods~\cite{Zhang2019, Zhao2021}, we employ a hybrid loss function to balance the outlier rejection task and the camera pose estimation task,
\begin{align}
\mathcal{L}=\mathcal{L}_{cls}+\tau\mathcal{L}_{ess},
\end{align}
where $\mathcal{L}_{cls}$ is a binary classification loss function, $\mathcal{L}_{ess}$ is a essential matrix loss, and $\tau$ is a weight factor. The $\mathcal{L}_{cls}$ can be denoted as,
\begin{align}
\mathcal{L}_{cls}=\sum_{j=0}^{M} H(w_j \odot o_j,y_j),
\end{align}
where $H(\cdot)$ is the binary cross entropy loss function, $w_j$ is an adaptive temperature. $o_j$ is the output weights of the $j$-th pruning module, $\odot$ denotes the Hadamard Product, and $y_j$ is the ground-truth labels.

The essential matrix loss $\mathcal{L}_{cls}$ can be denoted as a geometry loss,
\begin{align}
\mathcal{L}_{ess}=\frac{(\bm{p}'^{T}\hat{\bm{E}}\bm{p})^{2}}{\|\bm{Ep}\|_{[1]}^{2}+\|\bm{Ep}\|_{[2]}^{2}+\|\bm{Ep}'\|_{[1]}^{2}+\|\bm{Ep}'\|_{[2]}^{2}},
\end{align}
where $\bm{E}$ and $\hat{\bm{E}}$ denote the ground-truth essential matrix and the predicted essential matrix. $p$ and $p'$ are the virtual correspondence coordinates generated from E, $\|\cdot\|_j$ refers to the $j$-th element of vector.

\input{figures/camera_pose_yfcc}

\section{Experiments}
\label{sec:experiments}

\subsection{Experimental Settings}
\textbf{Datasets.}
To demonstrate the effectiveness of our proposed CorrMoE, we conduct experiments across various in-domain and out-of-domain scenes. For in-domain scenes, we conduct experiments on the YFCC100M~\cite{Thomee2016} dataset and the SUN3D~\cite{xiao2013sun3d} dataset. YFCC100M contains 68 sequences for training and validation purposes, and 4 sequences for testing, while SUN3D contains 239 sequences for training and validation purposes, and 15 for testing. For out-of-domain scenes, we evaluate methods trained on the YFCC100M dataset using the Zero Shot Evaluation Benchmark (ZEB) introduced by ~\cite{xuelun2024gim}, which is specifically designed to assess the cross-domain performance of methods. This benchmark is constructed by combining eight real-world datasets and four simulated datasets, covering diverse image resolutions, scene conditions, and viewpoints. Each dataset in this benchmark includes approximately 3,800 evaluation image pairs across five image overlap ratios (ranging from 10\% to 50\%).

\noindent\textbf{Evaluation Metrics.}
We evaluate the methods both quantitatively and qualitatively. For quantitative evaluation, to assess the accuracy of camera pose estimation, we calculate the Area Under the Curve (AUC) of pose errors at thresholds of 5°, 10°, and 20°. In addition, for the classification task of distinguishing inliers from outliers, we employ precision (P), recall (R), and F-score (F) as evaluation metrics. For qualitative evaluation, we apply methods to correspondence pruning and 3D reconstruction tasks, and present visual comparisons of different methods on both tasks.

\input{figures/camera_pose_sun3d}
\input{figures/outlier_rejection}

\noindent\textbf{Implementation Details.}
We use the SIFT detector~\cite{Lowe2004} to generate $N \times 4$ candidate correspondences, where $N = 2000$. The initial correspondences are subsequently pruned twice using the proposed module, each with a pruning rate of 0.5, thereby obtaining the $N/4$ reliable correspondence set. In PMix, $p_{mix}^{start}$ and $p_{mix}^{start}$ are set to 0.2 and 0.5, respectively. The number of neighbors for KNN is set to 9 and 6 in the first and second pruning modules, respectively. The number of clusters in the OA Filter is set to 250. The Bi-Fusion MoE is stacked four times, each with three experts, and the top-1 expert is selected in each layer. Our framework is implemented in PyTorch and trained on four NVIDIA A100 GPUs for 500k iterations. The Adam optimizer~\cite{kingma2014adam} is adopted with a batch size of 32 and an initial learning rate of 0.001, which is then gradually reduced for every 20k iterations with a decay factor of 0.9.

\begin{figure}[t]
\begin{center}
    \includegraphics[width=0.8\linewidth]{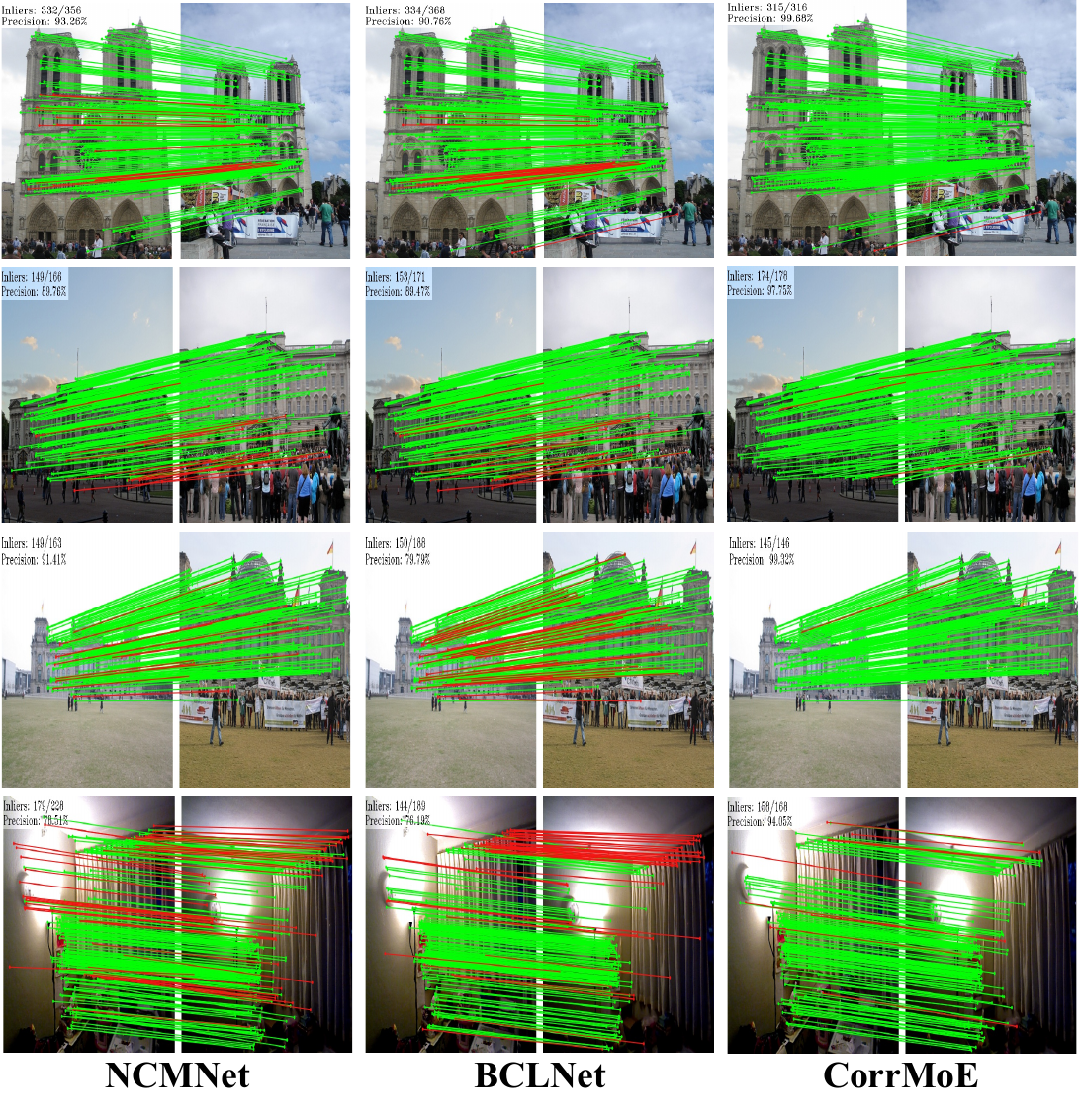}
    \caption{Visualization results on YFCC100M and SUN3D. From left to right are the results of NCMNet, BCLNet and our CorrMoE. The green lines represent inliers, the red lines represent outliers.}\label{fig:camera_pose_estimate}
    \vspace{0.2in}
\end{center}
\end{figure} 

\input{figures/cross_scene}

\subsection{Downstream Tasks}
\noindent\textbf{Camera Pose Estimation.}
Camera pose estimation determines the relative position and orientation between two cameras using identified inliers from the model. The results on the YFCC100M~\cite{Thomee2016} and SUN3D~\cite{xiao2013sun3d} datasets are shown in Table~\ref{tab:camera_pose_yfcc} and Table~\ref{tab:camera_pose_sun3d}. As we can see, our CorrMoE achieves the best performance in all scenes. For example, on the YFCC100M dataset, CorrMoE outperforms the recent SOTA method BCLNet by 2.28, 2.16, and 1.94 in terms of AUC@5°, AUC@10°, and AUC@20°, respectively. Moreover, the visualization results are shown in Fig.~\ref{fig:camera_pose_estimate}. When comparing existing advanced methods BCLNet and NCMNet, our proposed CorrMoE achieves the best visual performance.

\noindent\textbf{Outlier Rejection.}
Outlier rejection measures the model's ability to distinguish inliers from outliers. Correspondences are considered outliers if their epipolar distance exceeds the threshold of $10^{-4}$. Table~\ref{tab:outlier_rejection} compares the results on the YFCC100M dataset. Our method achieves the SOTA results in precision, recall, and F-score, surpassing the second-best method by 1.31, 0.31, and 0.57, respectively.

\begin{figure}[t]
\begin{center}
    \includegraphics[width=0.8\linewidth]{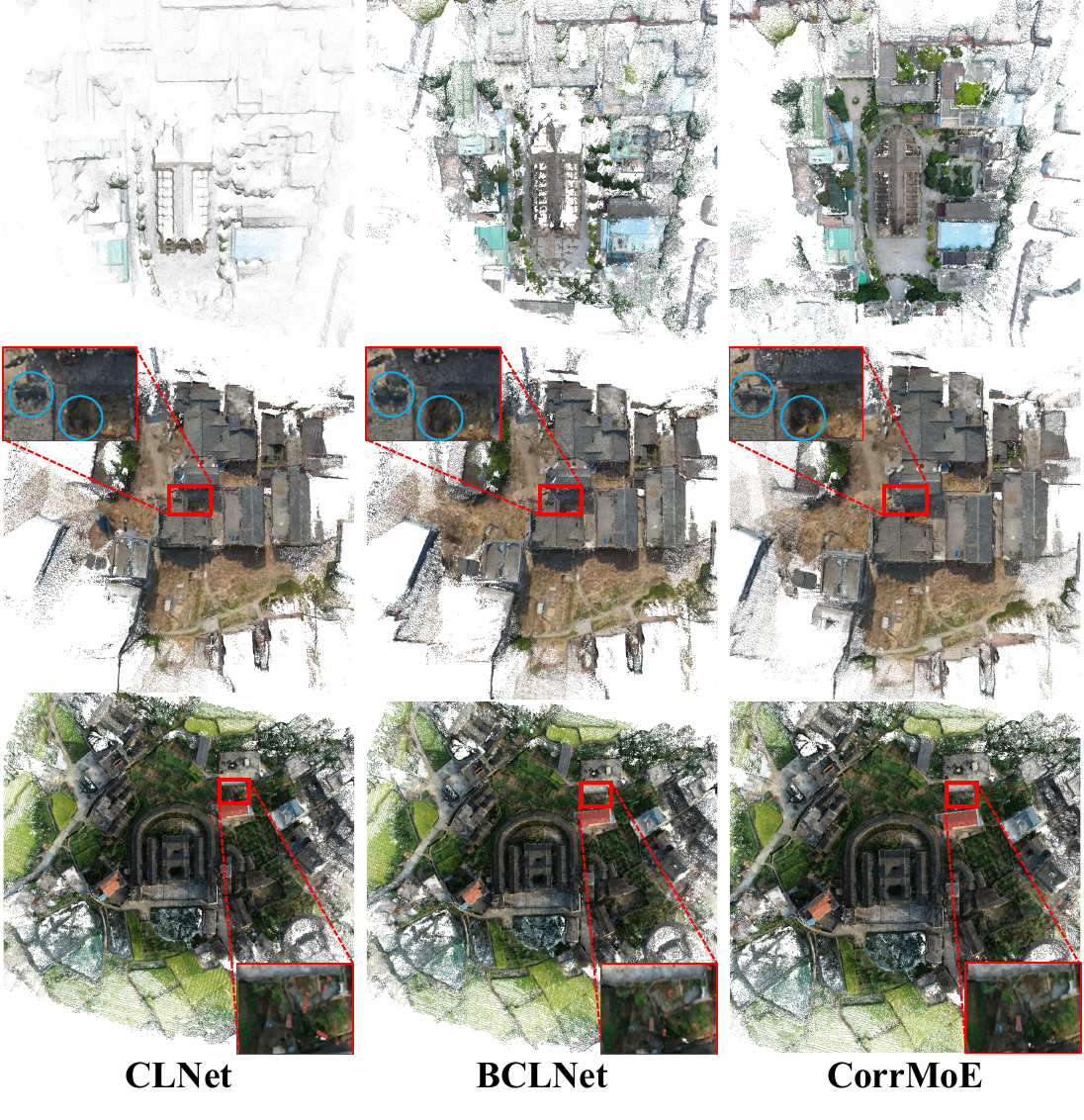}
    \caption{3D reconstruction results. From left to right are the results of CLNet, BCLNet and our CorrMoE.}\label{fig:3d_reconstruction}
    \vspace{0.1in}
\end{center}
\end{figure}

\noindent\textbf{3D Reconstruction.}
The quality of reconstructed 3D models depends heavily on the accuracy of the correspondence pruning model. Therefore, we apply our framework to 3D reconstruction tasks, using Structure-from-Motion (SfM)~\cite{ullman1979interpretation} for 3D model reconstruction from 2D correspondences and COLMAP~\cite{schonberger2016structure} for triangulation of the reference 3D models. As shown in Fig.~\ref{fig:3d_reconstruction}, our method preserves more fine-grained details, especially in regions with complex geometry or low texture. 
To be specific, in the first row, both CLNet and BCLNet fail to accurately reconstruct the building, while CorrMoE preserves its overall structure.
In the second row, CorrMoE handles shadow boundaries more effectively than CLNet and BCLNet, avoiding the blur and distortion introduced by other methods.
In the third row, CorrMoE produces fewer artifacts and noise during reconstruction. The results demonstrate the robustness of our pruning model in selecting high-quality inliers for reconstruction.

\input{figures/cross_domain}
\subsection{Cross-Scene and Cross-Domain Analysis}
\noindent\textbf{Cross-Scene Evaluation.}
To gain a deeper understanding of the methods' performance across both common and challenging scenes, we splits the test scenes from YFCC100M into four distinct sub-scenes: Buckingham Palace (BUC), Notre-Dame Front Facade (NOT), Reichstag (REI), and Sacré Coeur (SAC). While these sub-scenes belong to the same domain, each exhibits unique characteristics that challenge the methods' ability to adapt from training scenes to testing scenes. The AUC@5° and F-score are reported across four sub-scenes to underscore the performance variations induced by these scene-specific features.

As shown in Table~\ref{table:cross scene}, the proposed method consistently achieves superior performance across all four sub-scene. To be specific, it demonstrates substantial advantages in the BUC and SAC scenes, outperforming the second-best approaches by 2.55 and 5.52 in AUC@5°, respectively. The notable performance gains in targeted scenes suggest that the Bi-Fusion MoE architecture enhances the generalization to common and more challenging scenes that other existing methods struggle with.

\noindent\textbf{Cross-Domain Evaluation.}
To evaluate the generalization performance of the methods across out-of-domain scenes, we employ the zero-shot evaluation benchmark~\cite{xuelun2024gim}. This benchmark consists of 12 datasets, each containing completely different scenes, such as driving scenes, aerial scenes, sunlight changing scenes, season changing scenes, and more. We train all methods on the YFCC100M dataset and test them on the 12 cross-domain datasets separately. For simplicity, the abbreviations for these 12 datasets are derived from the first three letters of their names. 

The cross-domain evaluation results are shown in Table~\ref{tab:cross domain}. Our proposed method achieves a significant lead on most of the 12 datasets, with an average AUC@5° score of 22.80, surpassing the recent advanced method BCLNet by 8.38. Notably, on the ETO dataset, it outperforms the second-best method by 15.63, achieving a performance improvement of 75.51\%. The results demonstrate that by de-stylizing the training set input, our method achieves excellent generalization performance on most real and simulated datasets, greatly enhancing robustness and practicality in real-world applications.

\subsection{Ablation Study}
We conduct ablation studies on the YFCC100M and the zero-shot evaluation benchmark to validate the effectiveness of key components and parameters of the proposed method.

\input{figures/ablation_network}
\input{figures/ablation_stack}
\input{figures/ablation_pmix}

\noindent\textbf{Network Architecture.}
The proposed pruning module primarily consists of the De-stylization Dual Branch and the Bi-Fusion MoE, where the De-stylization Dual Branch includes both explicit and implicit branches. We validate the effectiveness of these components through a pruning strategy~\cite{Zhao2021}, with the performance on the YFCC100M measured by AUC@5° and AUC@20°. 

The results are shown in Table~\ref{tab:main_component}. Using only the Bi-Fusion MoE (Row1) or the dual-branch architecture 
(Row2) results in suboptimal performance, indicating that both proposed modules are indispensable. When combined with a single branch, the Bi-Fusion MoE yields improved results, with the explicit branch (Row 4) contributing more substantially than the implicit branch (Row 3), highlighting the stronger guidance provided by explicit representations. The full model (Row 5), which integrates the De-stylization Dual Branch with Bi-Fusion MoE, achieves the best performance, reaching an AUC@5° of 37.98 and an AUC@20° of 75.08, thereby validating the effectiveness of the proposed joint architecture.

\noindent\textbf{The stacking times of the Bi-Fusion MoE module.}
We evaluate the impact of stacking the Bi-Fusion MoE module a varying number of times on the YFCC100M dataset. Iterations Per Second (IPS) is used to assess the model’s efficiency. As shown in Table~\ref{tab:stack_times}, stacking the Bi-Fusion MoE module four times in our CorrMoE achieves the best balance between performance and efficiency.

\noindent\textbf{Impact of the PMix module on generalization.}
We replace the PMix module with Instance Normalization (IN)~\cite{ulyanov2016instance} and MixStyle~\cite{zhou2021domain}, both designed to capture different style characteristics. Then, we evaluate their generalization performance on YFCC100M and the zero-shot evaluation benchmark, which correspond to in-domain and cross-domain settings, respectively. As shown in Table~\ref{tab:pmix}, IN provides less generalization improvements compared to the other two. Although MixStyle and PMix show similar out-of-domain gains, PMix outperforms on the in-domain dataset. The results indicate that PMix not only improves cross-domain generalization but also accelerates model convergence, providing benefits in both training efficiency and robustness.

\section{Conclusion}
\label{sec:conclusion}

In this paper, we present CorrMoE, an innovative approach designed to tackle the challenges of correspondence pruning across varying scenes and domains. To improve cross-domain generalization, we introduce the De-stylization Dual Branch, which leverages Progressive Mixstyle to extract de-stylized local graph features implicitly and explicitly. Then, we propose the Bi-Fusion MoE module to adaptively integrate these features, where diverse experts focus on different spatial and geometric patterns, enabling effective handling of rare and complex scenes. Numerous experiments on multiple benchmarks show that our CorrMoE consistently achieves state-of-the-art performance, validating its effectiveness in improving generalization and its applicability to real-world visual correspondence tasks.

\section*{Acknowledgement}
This work is funded by Nanjing University-China Mobile Communications Group Co., Ltd. Joint Institute, and the National Natural Science Foundation of China (Grant No. 62372223 and U24A20330).  

\newpage
\bibliography{ref}

\end{document}

%% file: figures/camera_pose_yfcc.tex
\begin{table}[t]
\centering
\renewcommand{\arraystretch}{1.}
\caption{Comparison results on YFCC100M for camera pose estimation. AUC measures the results with different thresholds.}
\scalebox{1}{
\begin{tabular}{l|p{1.5cm}<{\centering}|p{1cm}<{\centering}p{1cm}<{\centering}p{1cm}<{\centering}} 
\midrule
\multirow{2}*{Method} & \multicolumn{3}{c}{Pose Estimation AUC $(\%)$}\\
\cmidrule{2-4}
 & @5$^\circ$ & @10$^\circ$ & @20$^\circ$ \\
\midrule
OANet~\cite{Zhang2019}     & 15.92 & 35.93 & 57.11  \\
CLNet~\cite{Zhao2021}       & 25.28 & 45.82 & 65.44  \\
TNet~\cite{zhong2021t}      & 20.53 & 42.65 & 63.20  \\
MS$^2$DGNet~\cite{Dai2022} & 20.61 & 42.90 & 64.26  \\
ConvMatch~\cite{convmatch2023}  & 26.83 & 49.14 & 67.91  \\
NCMNet~\cite{liu2023ncm}    & 34.51 & 55.34 & 72.40  \\
UMatch~\cite{li2024u}     & 30.84 & 52.04 & 69.65  \\
MGNet~\cite{dai2024mgnet}  & 32.32 & 53.40 & 71.59  \\
BCLNet~\cite{miao2024bclnet}   & 35.70 & 56.62 & 73.14  \\
DeMatch~\cite{zhang2024dematch}     & 30.91 & 52.71 & 70.35  \\
\midrule
\rowcolor{gray!25}
\textbf{CorrMoE~(Ours)} & \bf37.98 & \bf58.78 & \bf75.08 \\

\midrule
\end{tabular}
}
\label{tab:camera_pose_yfcc}
\end{table}

%% file: figures/camera_pose_sun3d.tex
\begin{table}[t]
\centering
\renewcommand{\arraystretch}{1.}
\caption{Comparison results on SUN3D for camera pose estimation. AUC measures the results with different thresholds.}
\scalebox{1}{
\begin{tabular}{l|p{1.5cm}<{\centering}|p{1cm}<{\centering}p{1cm}<{\centering}p{1cm}<{\centering}} 
\midrule
\multirow{2}*{Method} & \multicolumn{3}{c}{Pose Estimation AUC $(\%)$}\\
\cmidrule{2-4}
 & @5$^\circ$ & @10$^\circ$ & @20$^\circ$ \\
\midrule
PointCN~\cite{Yi2018}     & 3.05 & 10.00 & 24.06  \\
CLNet~\cite{Zhao2021}     & 5.88 & 16.39 & 32.96  \\
OANet~\cite{Zhang2019}     & 5.93 & 16.91 & 34.32  \\
MS$^2$DGNet~\cite{Dai2022} & 5.88 & 16.83 & 34.28  \\
LMCNet~\cite{convmatch2023}  & 7.08 & 19.09 & 37.15  \\
NCMNet~\cite{liu2023ncm}    & 8.13 & 20.42 & 38.17  \\
BCLNet~\cite{miao2024bclnet}   & 7.68 & 19.74 & 37.38  \\
\midrule
\rowcolor{gray!25}
\textbf{CorrMoE~(Ours)} & \bf8.18 & \bf{20.53} & \bf{38.30} \\

\midrule
\end{tabular}
}
\label{tab:camera_pose_sun3d}
\end{table}

%% file: figures/outlier_rejection.tex
\begin{table}[t]
\centering
\renewcommand{\arraystretch}{1.}
\caption{Comparison results on YFCC100M for outlier rejection. Precision, Recall and F-Score are reported.}
\scalebox{1}{
\begin{tabular}{l|p{1.3cm}<{\centering}p{1.3cm}<{\centering}p{1.3cm}<{\centering}} 
\midrule
\multirow{2}*{Method} & \multicolumn{3}{c}{Outlier Rejection} \\
\cmidrule{2-4}
 & Precision & Recall & F-Score \\
\midrule
OANet~\cite{Zhang2019}     & 68.05 & 68.41 & 68.23   \\
CLNet~\cite{Zhao2021}       & 75.05 & 76.41 & 75.72   \\
TNet~\cite{zhong2021t}      & 71.09 & 72.58 & 71.83   \\
MS$^2$DGNet~\cite{Dai2022} & 72.61 & 73.86 & 73.23   \\
ConvMatch~\cite{convmatch2023}  & 73.12 & 74.39 & 73.75   \\
NCMNet~\cite{liu2023ncm}    & 77.24 & 78.57 & 77.90   \\
UMatch~\cite{li2024u}      & 73.97 & 75.72 & 74.83   \\
MGNet~\cite{dai2024mgnet}   & 76.97 & 79.35 & 78.14   \\
BCLNet~\cite{miao2024bclnet}   & 77.40 & 79.82 & 78.59   \\
DeMatch~\cite{zhang2024dematch}   & 74.11 & 76.04 & 75.06   \\
\midrule
\rowcolor{gray!25}
\textbf{CorrMoE~(Ours)}   & \bf78.71 & \bf80.13 & \bf79.16  \\
\midrule
\end{tabular}
}
\label{tab:outlier_rejection}
\end{table}

%% file: figures/cross_scene.tex
\begin{table}[H]
\renewcommand\arraystretch{1.}
\centering
\setlength\tabcolsep{3pt}
    \caption{Cross-scene evaluation results. The four scenes are split from the YFCC100M test set. AUC@5$^\circ$ and F-score are reported.}
\resizebox{\linewidth}{!}{
\begin{tabular}{@{}l|c|c|c|c@{}}
    \toprule
    \multirow{2}{*}{Method} & BUC & NOT & REI & SAC \\
\cmidrule(lr){2-5}    & AUC@5$^\circ$ / F1 & AUC@5$^\circ$ / F1 & AUC@5$^\circ$ / F1 & AUC@5$^\circ$ / F1\\
    \midrule
    OANet~\cite{Zhang2019}  & 7.85 / 55.91 & 12.11 / 69.41 & 27.81 / 77.67 & 15.89 / 68.12 \\
    CLNet~\cite{Zhao2021}  & 16.46 / 65.68 & 17.47 / 76.65 & 32.93 / 80.59 & 33.66 / 78.86 \\
    TNet~\cite{zhong2021t} & 12.92 / 61.20 & 19.74 / 74.13 & 32.83 / 79.52 & 16.60 / 71.05 \\
    MS$^2$DGNet~\cite{Dai2022} & 14.08 / 64.12 & 15.42 / 73.74 & 32.74 / 79.16 & 20.39 / 74.28 \\
    ConvMatch~\cite{convmatch2023} & 16.02 / 63.99 & 23.03 / 75.24 & 37.36 / 79.82 & 30.16 / 74.58 \\
    NCMNet~\cite{liu2023ncm}  & 25.61 / 68.61 & 26.32 / 79.39 & 43.05 / 81.95 & 44.05 / 81.32 \\
    UMatch~\cite{li2024u} & 18.36 / 64.54 & 23.57 / 75.88 & 39.69 / 80.86 & 41.22 / 77.02 \\
    MGNet~\cite{dai2024mgnet}  & 20.29 / 68.94 & 21.54 / 78.79 & 43.63 / 82.15 & 43.48 / 81.47 \\
    BCLNet~\cite{miao2024bclnet}  & 25.76 / 69.46 & 26.32 / 79.08 & 45.35 / 82.66 & 45.23 / 82.05 \\
    DeMatch~\cite{zhang2024dematch} & 17.04 / 64.48 & 25.81 / 76.72 & 42.18 / 80.86 & 38.14 / 77.08 \\
    \midrule
    \rowcolor{gray!25}
    \textbf{CorrMoE~(Ours)} & \textbf{28.31 / 70.11} & \textbf{27.40 / 80.56} & \textbf{45.58 / 82.53} & \textbf{50.75 / 83.44} \\
    \bottomrule
\end{tabular}
}
\label{table:cross scene}
\end{table}

%% file: figures/cross_domain.tex
\begin{table*}[t]
\renewcommand\arraystretch{1.1}
\centering
    \caption{Cross-domain evaluation results on the zero-shot evaluation benchmark. AUC@5$^\circ$ is reported. The proposed method achieves leading performance on most of the 12 datasets.}
\resizebox{\textwidth}{!}{%
\begin{tabular}{@{}l|cccccccc|cccc|c}
    \toprule
    \multirow{2}{*}{Method} & \multicolumn{8}{c}{\cellcolor{blue!20}\textit{Real}} & \multicolumn{4}{c}{\cellcolor{red!20}\textit{Simulate}} & Mean            \\
        & GL3   & BLE   & ETI   & ETO   & KIT   & WEA   & SEA  & NIG & MUL & SCE & ICL & GTA & AUC@5$^\circ(\%)$\\
    \midrule
    CLNet~\cite{Zhao2021}        & 15.52   & 13.76    & 11.14   & 16.75    & 21.04   & 17.90  & 27.69 & 12.08   & 10.73  & 3.27   & 5.77  & 11.61  & 13.94     \\
    ConvMatch~\cite{convmatch2023}
    & 11.83   & 11.57    & 10.35   & 12.36    & 22.80   & 17.12  & 28.23 & 11.99   & \textbf{20.64}  & 4.88   & 6.21  & 9.90  & 13.99     \\
    UMatch~\cite{li2024u} 
    & 8.00    & 7.92     & 4.93    & 4.91     & 17.01   & 11.38  & 18.96 & 9.05   & 8.72  & 2.51   & 3.29  & 5.38  & 8.51      \\
    NCMNet~\cite{liu2023ncm} 
    & 14.17   & 14.43    & 10.38   & 13.23    & 23.92   & 19.08  &  29.74 & 12.82   & 8.11  & 4.14   & 7.51  & 6.72  & 13.69     \\
    MGNet~\cite{dai2024mgnet}
    & 14.40   & 13.58    & 10.25   & 12.14    & 21.21   & 17.27  & 26.41 & 12.05  & 13.51  & 3.70  & 6.48  & 10.99  & 13.50     \\
    BCLNet~\cite{miao2024bclnet}
    & 17.00   & 15.23    & 12.84   & 15.51    & 23.26   & 18.66  & 28.17 & 11.94   & 7.83  & 4.64   & 7.48  & 10.03  & 14.42     \\
    PGFNet~\cite{liu2023pgfnet}
    & 12.80   & 11.30    & 17.37   & 20.70    & 22.65   & 16.07  & 24.96 & 11.33   & 14.13  & 2.98   & 4.64  & 12.10  & 14.25     \\
    \midrule
    \rowcolor{gray!25}
    \textbf{CorrMoE~(Ours)}          & \textbf{28.65}   & \textbf{22.34}    & \textbf{30.48}   & \textbf{36.33}    & \textbf{26.24}   & \textbf{21.21}  & \textbf{32.80} & \textbf{14.08}   & 18.19  & \textbf{6.45}   & \textbf{12.43}  & \textbf{24.37}  & \textbf{22.80}    \\
    \bottomrule
\end{tabular}
}
\vspace{0.1in}
\label{tab:cross domain}
\end{table*}

%% file: figures/ablation_network.tex
\begin{table}[t]
\centering
\caption{Ablation study for the main components of CorrMoE.}
\begin{tabular}{ccc|cc}
\toprule
\multirow{2}{*}{Bi-Fusion} & \multirow{2}{*}{Implicit} & \multirow{2}{*}{Explicit} & \multicolumn{2}{c}{AUC $(\%)$} \\
\cmidrule(lr){4-5}
MoE & Branch & Branch & @5$^\circ$ & @20$^\circ$ \\
\midrule
$\surd$ &  & & 34.04 & 72.28 \\
 & $\surd$ & $\surd$ & 34.61 & 72.76 \\
$\surd$ & $\surd$ & & 35.17 & 72.62 \\
$\surd$ & & $\surd$ & 37.01 & 74.22 \\
$\surd$ & $\surd$ & $\surd$ & \textbf{37.98} & \textbf{75.08} \\
\bottomrule
\end{tabular}
\label{tab:main_component}
\end{table}

%% file: figures/ablation_stack.tex
\begin{table}[t]
\centering
\renewcommand{\arraystretch}{1.}
\caption{Ablation study of stacking times for the Bi-Fusion MoE module in our CorrMoE. IPS denotes Iterations Per Second.}
\begin{tabular}{c|ccc}
\toprule
Stack Times & IPS $\uparrow$ & AUC@5$^\circ$ & AUC@20$^\circ$ \\
\midrule
3 & \textbf{10.11} & 37.47 & 74.66 \\
\textbf{4} & 8.89 & \textbf{37.98} & \textbf{75.08} \\
5 & 7.12 & 36.43 & 73.85 \\
\bottomrule
\end{tabular}
\label{tab:stack_times}
\end{table}

%% file: figures/ablation_pmix.tex
\begin{table}[t]
\centering
\renewcommand{\arraystretch}{1.1}
\caption{Ablation study of PMix's impact on model generalization, where IN denotes Instance Normalization.}
\begin{tabular}{c|c|cc}
\toprule
\multirow{2}{*}{Module} & \multicolumn{1}{c|}{Cross-Domain} & \multicolumn{2}{c}{In-Domain} \\
\cmidrule(lr){2-4}
 & AUC@5$^\circ$ & AUC@5$^\circ$ & AUC@20$^\circ$ \\
\midrule
IN~\cite{ulyanov2016instance} & 21.67 & 35.34 & 73.53 \\
Mixstyle~\cite{zhou2021domain} & 22.39 & 37.02 & 74.25 \\
\textbf{PMix (Ours)} & \textbf{22.80} & \textbf{37.98} & \textbf{75.08} \\
\bottomrule
\end{tabular}
\label{tab:pmix}
\end{table}

%% file: main.bbl
\begin{thebibliography}{47}
\providecommand{\natexlab}[1]{#1}
\providecommand{\url}[1]{\texttt{#1}}
\expandafter\ifx\csname urlstyle\endcsname\relax
  \providecommand{\doi}[1]{doi: #1}\else
  \providecommand{\doi}{doi: \begingroup \urlstyle{rm}\Url}\fi

\bibitem[Agarwal et~al.(2011)Agarwal, Furukawa, Snavely, Simon, Curless, Seitz, and Szeliski]{agarwal2011building}
S.~Agarwal, Y.~Furukawa, N.~Snavely, I.~Simon, B.~Curless, S.~M. Seitz, and R.~Szeliski.
\newblock Building rome in a day.
\newblock \emph{Communications of the ACM}, 54\penalty0 (10):\penalty0 105--112, 2011.

\bibitem[Barath et~al.(2019)Barath, Matas, and Noskova]{Barath2019}
D.~Barath, J.~Matas, and J.~Noskova.
\newblock Magsac: marginalizing sample consensus.
\newblock In \emph{Proceedings of the IEEE/CVF Conference on Computer Vision and Pattern Recognition}, pages 10197--10205, 2019.

\bibitem[Campos et~al.(2021)Campos, Elvira, Rodr{\'\i}guez, Montiel, and Tard{\'o}s]{campos2021orb}
C.~Campos, R.~Elvira, J.~J.~G. Rodr{\'\i}guez, J.~M. Montiel, and J.~D. Tard{\'o}s.
\newblock Orb-slam3: An accurate open-source library for visual, visual--inertial, and multimap slam.
\newblock \emph{IEEE Transactions on Robotics}, 37\penalty0 (6):\penalty0 1874--1890, 2021.

\bibitem[Chum and Matas(2005)]{Chum2005a}
O.~Chum and J.~Matas.
\newblock Matching with prosac-progressive sample consensus.
\newblock In \emph{Proceedings of the IEEE/CVF Conference on Computer Vision and Pattern Recognition}, pages 220--226. IEEE, 2005.

\bibitem[Dai et~al.(2022)Dai, Liu, Ma, Wei, Lai, Yang, and Chen]{Dai2022}
L.~Dai, Y.~Liu, J.~Ma, L.~Wei, T.~Lai, C.~Yang, and R.~Chen.
\newblock Ms2dg-net: Progressive correspondence learning via multiple sparse semantics dynamic graph.
\newblock In \emph{Proceedings of the IEEE/CVF Conference on Computer Vision and Pattern Recognition}, pages 8973--8982, 2022.

\bibitem[Dai et~al.(2024)Dai, Du, Zhang, and Tang]{dai2024mgnet}
L.~Dai, X.~Du, H.~Zhang, and J.~Tang.
\newblock Mgnet: Learning correspondences via multiple graphs.
\newblock In \emph{Proceedings of the AAAI Conference on Artificial Intelligence}, pages 3945--3953, 2024.

\bibitem[DeTone et~al.(2018)DeTone, Malisiewicz, and Rabinovich]{DeTone2018}
D.~DeTone, T.~Malisiewicz, and A.~Rabinovich.
\newblock Superpoint: Self-supervised interest point detection and description.
\newblock In \emph{Proceedings of the IEEE/CVF Conference on Computer Vision and Pattern Recognition}, pages 224--236, 2018.

\bibitem[Eigen et~al.(2013)Eigen, Ranzato, and Sutskever]{eigen2013learning}
D.~Eigen, M.~Ranzato, and I.~Sutskever.
\newblock Learning factored representations in a deep mixture of experts.
\newblock \emph{arXiv preprint arXiv:1312.4314}, 2013.

\bibitem[Fedus et~al.(2022)Fedus, Zoph, and Shazeer]{fedus2022switch}
W.~Fedus, B.~Zoph, and N.~Shazeer.
\newblock Switch transformers: Scaling to trillion parameter models with simple and efficient sparsity.
\newblock \emph{Journal of Machine Learning Research}, 23\penalty0 (120):\penalty0 1--39, 2022.

\bibitem[Fischler and Bolles(1981)]{Fischler1981}
M.~A. Fischler and R.~C. Bolles.
\newblock Random sample consensus: a paradigm for model fitting with applications to image analysis and automated cartography.
\newblock \emph{Communications of the ACM}, 24\penalty0 (6):\penalty0 381--395, 1981.

\bibitem[Gross et~al.(2017)Gross, Ranzato, and Szlam]{gross2017hard}
S.~Gross, M.~Ranzato, and A.~Szlam.
\newblock Hard mixtures of experts for large scale weakly supervised vision.
\newblock In \emph{Proceedings of the IEEE/CVF Conference on Computer Vision and Pattern Recognition}, pages 6865--6873, 2017.

\bibitem[Jordan and Jacobs(1994)]{jordan1994hierarchical}
M.~I. Jordan and R.~A. Jacobs.
\newblock Hierarchical mixtures of experts and the em algorithm.
\newblock \emph{Neural Computation}, 6\penalty0 (2):\penalty0 181--214, 1994.

\bibitem[Kerbl et~al.(2023)Kerbl, Kopanas, Leimk{\"u}hler, and Drettakis]{kerbl3Dgaussians}
B.~Kerbl, G.~Kopanas, T.~Leimk{\"u}hler, and G.~Drettakis.
\newblock 3d gaussian splatting for real-time radiance field rendering.
\newblock \emph{ACM Transactions on Graphics}, 42\penalty0 (4), July 2023.

\bibitem[Kingma and Ba(2014)]{kingma2014adam}
D.~P. Kingma and J.~L. Ba.
\newblock Adam: A method for stochastic optimization.
\newblock In \emph{Proceedings of the International Conference on Learning Representations}, pages 1--15, 2014.

\bibitem[Lepikhin et~al.(2020)Lepikhin, Lee, Xu, Chen, Firat, Huang, Krikun, Shazeer, and Chen]{lepikhin2020gshard}
D.~Lepikhin, H.~Lee, Y.~Xu, D.~Chen, O.~Firat, Y.~Huang, M.~Krikun, N.~Shazeer, and Z.~Chen.
\newblock Gshard: Scaling giant models with conditional computation and automatic sharding.
\newblock \emph{arXiv preprint arXiv:2006.16668}, 2020.

\bibitem[Li et~al.(2024)Li, Zhang, and Ma]{li2024u}
Z.~Li, S.~Zhang, and J.~Ma.
\newblock U-match: Exploring hierarchy-aware local context for two-view correspondence learning.
\newblock \emph{IEEE Transactions on Pattern Analysis and Machine Intelligence}, 2024.

\bibitem[Liao et~al.(2023)Liao, Zhang, Xu, Shi, and Xiao]{liao2023sga}
T.~Liao, X.~Zhang, Y.~Xu, Z.~Shi, and G.~Xiao.
\newblock Sga-net: A sparse graph attention network for two-view correspondence learning.
\newblock \emph{IEEE Transactions on Circuits and Systems for Video Technology}, 33\penalty0 (12):\penalty0 7578--7590, 2023.

\bibitem[Liao et~al.(2024)Liao, Zhang, Zhao, Wang, and Xiao]{liao2023vsformer}
T.~Liao, X.~Zhang, L.~Zhao, T.~Wang, and G.~Xiao.
\newblock Vsformer: Visual-spatial fusion transformer for correspondence pruning.
\newblock In \emph{Proceedings of the AAAI Conference on Artificial Intelligence}, 2024.

\bibitem[Liu and Yang(2023)]{liu2023ncm}
X.~Liu and J.~Yang.
\newblock Progressive neighbor consistency mining for correspondence pruning.
\newblock In \emph{Proceedings of the IEEE/CVF Conference on Computer Vision and Pattern Recognition}, pages 9527--9537, 2023.

\bibitem[Liu et~al.(2023)Liu, Xiao, Chen, and Ma]{liu2023pgfnet}
X.~Liu, G.~Xiao, R.~Chen, and J.~Ma.
\newblock Pgfnet: Preference-guided filtering network for two-view correspondence learning.
\newblock \emph{IEEE Transactions on Image Processing}, 32:\penalty0 1367--1378, 2023.

\bibitem[Lowe(2004)]{Lowe2004}
D.~G. Lowe.
\newblock Distinctive image features from scale-invariant keypoints.
\newblock \emph{International Journal of Computer Vision}, 60\penalty0 (2):\penalty0 91--110, 2004.

\bibitem[Ma et~al.(2025)Ma, Zhuang, Hao, and King]{ma20253d}
Y.~Ma, Y.~Zhuang, J.~Hao, and I.~King.
\newblock 3d-moe: A mixture-of-experts multi-modal llm for 3d vision and pose diffusion via rectified flow.
\newblock \emph{arXiv preprint arXiv:2501.16698}, 2025.

\bibitem[Masoudnia and Ebrahimpour(2014)]{masoudnia2014mixture}
S.~Masoudnia and R.~Ebrahimpour.
\newblock Mixture of experts: a literature survey.
\newblock \emph{Artificial Intelligence Review}, 42:\penalty0 275--293, 2014.

\bibitem[Miao et~al.(2024)Miao, Xiao, Wang, and Yu]{miao2024bclnet}
X.~Miao, G.~Xiao, S.~Wang, and J.~Yu.
\newblock Bclnet: Bilateral consensus learning for two-view correspondence pruning.
\newblock In \emph{Proceedings of the AAAI Conference on Artificial Intelligence}, pages 4225--4232, 2024.

\bibitem[Mur-Artal and Tard{\'o}s(2017)]{mur2017orb}
R.~Mur-Artal and J.~D. Tard{\'o}s.
\newblock Orb-slam2: An open-source slam system for monocular, stereo, and rgb-d cameras.
\newblock \emph{IEEE Transactions on Robotics}, 33\penalty0 (5):\penalty0 1255--1262, 2017.

\bibitem[Mur-Artal et~al.(2015)Mur-Artal, Montiel, and Tardos]{mur2015orb}
R.~Mur-Artal, J.~M.~M. Montiel, and J.~D. Tardos.
\newblock Orb-slam: a versatile and accurate monocular slam system.
\newblock \emph{IEEE Transactions on Robotics}, 31\penalty0 (5):\penalty0 1147--1163, 2015.

\bibitem[Raguram et~al.(2012)Raguram, Chum, Pollefeys, Matas, and Frahm]{Raguram2012}
R.~Raguram, O.~Chum, M.~Pollefeys, J.~Matas, and J.-M. Frahm.
\newblock Usac: A universal framework for random sample consensus.
\newblock \emph{IEEE Transactions on Pattern Analysis and Machine Intelligence}, 35\penalty0 (8):\penalty0 2022--2038, 2012.

\bibitem[Sarlin et~al.(2021)Sarlin, Unagar, Larsson, Germain, Toft, Larsson, Pollefeys, Lepetit, Hammarstrand, Kahl, et~al.]{sarlin2021back}
P.-E. Sarlin, A.~Unagar, M.~Larsson, H.~Germain, C.~Toft, V.~Larsson, M.~Pollefeys, V.~Lepetit, L.~Hammarstrand, F.~Kahl, et~al.
\newblock Back to the feature: Learning robust camera localization from pixels to pose.
\newblock In \emph{Proceedings of the IEEE/CVF Conference on Computer Vision and Pattern Recognition}, pages 3247--3257, 2021.

\bibitem[Schonberger and Frahm(2016)]{schonberger2016structure}
J.~L. Schonberger and J.-M. Frahm.
\newblock Structure-from-motion revisited.
\newblock In \emph{Proceedings of the IEEE/CVF Conference on Computer Vision and Pattern Recognition}, pages 4104--4113, 2016.

\bibitem[Shazeer et~al.(2017)Shazeer, Mirhoseini, Maziarz, Davis, Le, Hinton, and Dean]{shazeer2017outrageously}
N.~Shazeer, A.~Mirhoseini, K.~Maziarz, A.~Davis, Q.~Le, G.~Hinton, and J.~Dean.
\newblock Outrageously large neural networks: The sparsely-gated mixture-of-experts layer.
\newblock \emph{arXiv preprint arXiv:1701.06538}, 2017.

\bibitem[Shen et~al.(2024)Shen, Cai, Yin, Müller, Li, Wang, Chen, and Wang]{xuelun2024gim}
X.~Shen, Z.~Cai, W.~Yin, M.~Müller, Z.~Li, K.~Wang, X.~Chen, and C.~Wang.
\newblock Gim: Learning generalizable image matcher from internet videos.
\newblock In \emph{Proceedings of the International Conference on Learning Representations}, 2024.

\bibitem[Thomee et~al.(2016)Thomee, Shamma, Friedland, Elizalde, Ni, Poland, Borth, and Li]{Thomee2016}
B.~Thomee, D.~A. Shamma, G.~Friedland, B.~Elizalde, K.~Ni, D.~Poland, D.~Borth, and L.-J. Li.
\newblock Yfcc100m: The new data in multimedia research.
\newblock \emph{Communications of the ACM}, 59\penalty0 (2):\penalty0 64--73, 2016.

\bibitem[Torr and Zisserman(2000)]{torr2000mlesac}
P.~H. Torr and A.~Zisserman.
\newblock Mlesac: A new robust estimator with application to estimating image geometry.
\newblock \emph{Computer Vision and Image Understanding}, 78\penalty0 (1):\penalty0 138--156, 2000.

\bibitem[Ullman(1979)]{ullman1979interpretation}
S.~Ullman.
\newblock The interpretation of structure from motion.
\newblock \emph{Proceedings of the Royal Society of London. Series B. Biological Sciences}, 203\penalty0 (1153):\penalty0 405--426, 1979.

\bibitem[Ulyanov et~al.(2016)Ulyanov, Vedaldi, and Lempitsky]{ulyanov2016instance}
D.~Ulyanov, A.~Vedaldi, and V.~Lempitsky.
\newblock Instance normalization: The missing ingredient for fast stylization.
\newblock \emph{arXiv preprint arXiv:1607.08022}, 2016.

\bibitem[Wu et~al.(2022)Wu, Wu, Xu, Wang, and Long]{wu2022flowformer}
H.~Wu, J.~Wu, J.~Xu, J.~Wang, and M.~Long.
\newblock Flowformer: Linearizing transformers with conservation flows.
\newblock In \emph{Proceedings of the International Conference on Machine Learning}, 2022.

\bibitem[Xiao et~al.(2013)Xiao, Owens, and Torralba]{xiao2013sun3d}
J.~Xiao, A.~Owens, and A.~Torralba.
\newblock Sun3d: A database of big spaces reconstructed using sfm and object labels.
\newblock In \emph{Proceedings of the IEEE International Conference on Computer Vision}, pages 1625--1632, 2013.

\bibitem[Yao et~al.(2021)Yao, Jia, Di, Li, and Wu]{yao2021decomposition}
C.~Yao, Y.~Jia, H.~Di, P.~Li, and Y.~Wu.
\newblock A decomposition model for stereo matching.
\newblock In \emph{Proceedings of the IEEE/CVF Conference on Computer Vision and Pattern Recognition}, pages 6091--6100, 2021.

\bibitem[Yi et~al.(2018)Yi, Trulls, Ono, Lepetit, Salzmann, and Fua]{Yi2018}
K.~M. Yi, E.~Trulls, Y.~Ono, V.~Lepetit, M.~Salzmann, and P.~Fua.
\newblock Learning to find good correspondences.
\newblock In \emph{Proceedings of the IEEE/CVF Conference on Computer Vision and Pattern Recognition}, pages 2666--2674, 2018.

\bibitem[Zhang et~al.(2019)Zhang, Sun, Luo, Yao, Zhou, Shen, Chen, Quan, and Liao]{Zhang2019}
J.~Zhang, D.~Sun, Z.~Luo, A.~Yao, L.~Zhou, T.~Shen, Y.~Chen, L.~Quan, and H.~Liao.
\newblock Learning two-view correspondences and geometry using order-aware network.
\newblock In \emph{Proceedings of the IEEE/CVF Conference on Computer Vision and Pattern Recognition}, pages 5845--5854, 2019.

\bibitem[Zhang and Ma(2023)]{convmatch2023}
S.~Zhang and J.~Ma.
\newblock Convmatch: Rethinking network design for two-view correspondence learning.
\newblock \emph{IEEE Transactions on Pattern Analysis and Machine Intelligence}, 2023.

\bibitem[Zhang et~al.(2024)Zhang, Li, Gao, and Ma]{zhang2024dematch}
S.~Zhang, Z.~Li, Y.~Gao, and J.~Ma.
\newblock Dematch: Deep decomposition of motion field for two-view correspondence learning.
\newblock In \emph{Proceedings of the IEEE/CVF Conference on Computer Vision and Pattern Recognition}, pages 20278--20287, 2024.

\bibitem[Zhao et~al.(2021)Zhao, Ge, Zhu, Zhao, Li, and Salzmann]{Zhao2021}
C.~Zhao, Y.~Ge, F.~Zhu, R.~Zhao, H.~Li, and M.~Salzmann.
\newblock Progressive correspondence pruning by consensus learning.
\newblock In \emph{Proceedings of the IEEE/CVF Conference on Computer Vision and Pattern Recognition}, pages 6464--6473, 2021.

\bibitem[Zhong et~al.(2021)Zhong, Xiao, Zheng, Lu, and Ma]{zhong2021t}
Z.~Zhong, G.~Xiao, L.~Zheng, Y.~Lu, and J.~Ma.
\newblock T-net: Effective permutation-equivariant network for two-view correspondence learning.
\newblock In \emph{Proceedings of the IEEE/CVF International Conference on Computer Vision}, pages 1950--1959, 2021.

\bibitem[Zhou et~al.(2021)Zhou, Yang, Qiao, and Xiang]{zhou2021domain}
K.~Zhou, Y.~Yang, Y.~Qiao, and T.~Xiang.
\newblock Domain generalization with mixstyle.
\newblock \emph{arXiv preprint arXiv:2104.02008}, 2021.

\bibitem[Zhu et~al.(2024{\natexlab{a}})Zhu, Sun, Cao, and Hu]{zhu2024task}
P.~Zhu, Y.~Sun, B.~Cao, and Q.~Hu.
\newblock Task-customized mixture of adapters for general image fusion.
\newblock In \emph{Proceedings of the IEEE/CVF Conference on Computer Vision and Pattern Recognition}, pages 7099--7108, 2024{\natexlab{a}}.

\bibitem[Zhu et~al.(2024{\natexlab{b}})Zhu, Liu, He, Liao, Zheng, Xu, Wang, and Lu]{zhu2024corradaptor}
W.~Zhu, Y.~Liu, Y.~He, T.~Liao, K.~Zheng, X.~Xu, T.~Wang, and T.~Lu.
\newblock Corradaptor: Adaptive local context learning for correspondence pruning.
\newblock In \emph{Proceedings of the 27th European Conference on Artificial Intelligence}, 2024{\natexlab{b}}.

\end{thebibliography}
